\documentclass[sigconf]{acmart}

\AtBeginDocument{%
  \providecommand\BibTeX{{%
    \normalfont B\kern-0.5em{\scshape i\kern-0.25em b}\kern-0.8em\TeX}}}

\usepackage{booktabs} 
\usepackage{graphicx} 
\usepackage{subcaption}
\usepackage[utf8]{inputenc} 
\usepackage[T1]{fontenc}    
\usepackage{hyperref}       
\usepackage{url}            
\usepackage{booktabs}       
\usepackage{amsfonts}       
\usepackage{nicefrac}       
\usepackage{microtype}      
\usepackage{amsthm}
\usepackage{algorithm,algpseudocode}
\usepackage{amsthm}
\usepackage{verbatimbox}
\usepackage{accents}
\usepackage{graphicx}
\usepackage{ulem}
\usepackage{graphicx}
\usepackage{geometry}
\usepackage{microtype}
\usepackage{graphicx}
\usepackage{booktabs} 
\usepackage[font=small,labelfont=bf]{caption}
\usepackage{algorithm,algpseudocode}

\input{Definitions}

\begin{document}

\copyrightyear{2020} 
\acmYear{2020} 
\setcopyright{acmcopyright}\acmConference[CIKM '20]{Proceedings of the 29th ACM International Conference on Information and Knowledge Management}{October 19--23, 2020}{Virtual Event, Ireland}
\acmBooktitle{Proceedings of the 29th ACM International Conference on Information and Knowledge Management (CIKM '20), October 19--23, 2020, Virtual Event, Ireland}
\acmPrice{15.00}
\acmDOI{10.1145/3340531.3412705}
\acmISBN{978-1-4503-6859-9/20/10}




\fancyhead{}
\settopmatter{printacmref=false, printfolios=false}

\title{LiFT: A Scalable Framework for Measuring Fairness in ML Applications}

\author{Sriram Vasudevan}
\affiliation{
  \institution{LinkedIn Corporation}
}
\email{svasudevan@linkedin.com}

\author{Krishnaram Kenthapadi$^{1}$}\thanks{$^1$Work done while at LinkedIn}
\affiliation{
  \institution{Amazon AWS AI}
}
\email{kenthk@amazon.com}

\renewcommand{\shortauthors}{Vasudevan, Kenthapadi.}

\begin{abstract}
Many internet applications are powered by machine learned models, which are usually trained on labeled datasets obtained through either implicit / explicit user feedback signals or human judgments. Since societal biases may be present in the generation of such datasets, it is possible for the trained models to be biased, thereby resulting in potential discrimination and harms for disadvantaged groups. Motivated by the need for understanding and addressing algorithmic bias in web-scale ML systems and the limitations of existing fairness toolkits, we present the LinkedIn Fairness Toolkit (LiFT), a framework for scalable computation of fairness metrics as part of large ML systems. We highlight the key requirements in deployed settings, and present the design of our fairness measurement system. We discuss the challenges encountered in incorporating fairness tools in practice and the lessons learned during deployment at LinkedIn. Finally, we provide open problems based on practical experience.
\end{abstract}

\maketitle

\section{Introduction}\label{sec:intro}
Several large-scale internet applications make use of search and recommendation systems that are powered by algorithmic models and techniques.
Recent studies have shown that results produced by a biased machine learning model can result in discrimination and potential harms for disadvantaged groups~\cite{nips_2017_tutorial,crawford2017trouble}. A key reason is that machine learned models that are trained on data affected by societal biases may learn to act in accordance with them.

While there have been several efforts to build fairness toolkits that offer a comprehensive set of fairness metrics that can be measured, provide a suite of bias mitigation techniques, or enable the comparison of fairness metrics across various algorithms~\cite{adebayo2016fairml, agarwal2018reductions, bantilan2018themis, bellamy2018ai, friedler2019comparative, galhotra2017fairness, auditAI, saleiro2018aequitas, tramer2015fairtest, fairnessMeasures}, our experience at LinkedIn suggests that none of them addresses the need for tackling these problems at scale or integrating easily with ML pipelines (\S\ref{sec:relatedwork}). Motivated by the need for understanding and addressing algorithmic bias in web-scale ML systems, we present the LinkedIn Fairness Toolkit (LiFT), a framework for scalable computation of fairness metrics as part of large ML systems. We highlight the key requirements in deployed settings (\S\ref{sec:requirements}), and present the design and architecture of our system (\S\ref{sec:fairnessSystemArch} and \S\ref{sec:metrics}). We present results from deployments at LinkedIn and discuss the challenges encountered and lessons learned during deployment of fairness tools in practice (\S\ref{sec:results} and \S\ref{sec:impact}). Finally, we provide open problems and research directions based on our experiences (\S\ref{sec:conclusion}).

The key contributions of our work are as follows:
\begin{itemize}
\item Architecture of how bias measurement and mitigation tools can be integrated with production ML systems, to ensure monitoring and mitigation at each stage of the ML lifecycle.
\item Design of a fairness solution that scales to handle large datasets, and is flexible to use both in model training / scoring pipelines as well as for ad hoc data analysis.
\item Implementation of a fairness toolkit that measures biases in training data, computes fairness metrics for trained models, and detects {\em statistically significant} differences in model performance across different subgroups.
\item Discussion of challenges encountered and lessons learned during deployment as well as open problems based on practical experience.
\end{itemize}

\section{Problem Setting}\label{sec:requirements}
We present the key requirements for the adoption of fairness tools in practice as part of web-scale ML systems. For deployment in such systems, our experience suggests that bias measurement and mitigation solutions must be:

\begin{itemize}
\item {\bf Flexible}: Fairness tools should be usable as libraries for ad-hoc exploratory analyses (e.g., with Jupyter notebooks) as well as be conducive to deployment in production ML workflows that run on a regular cadence. It should also be straightforward to integrate these solutions with existing ML platforms, to increase the likelihood and ease of adoption by model developers.

\item {\bf Scalable}: Computation associated with bias measurement and mitigation should be amenable to being performed over several nodes in a distributed computing environment, since data parallelism enables fast computation over large datasets. Furthermore, since datasets are typically stored in distributed file systems and large-scale deployed ML model training workflows are typically setup in this fashion, ensuring that bias measurement and mitigation frameworks are distributed allows for easy integration and adoption.
\end{itemize}

Thus, our problem can be formulated as: \textit{Provide an architecture for integrating bias measurement and bias mitigation into production ML systems that operate on datasets stored in distributed file systems, and prescribe a design for fairness toolkits that are flexible to use, integrate easily with existing ML platforms, and scale to large datasets.}
\begin{figure*}[htbp]
	\centering
 	\vspace{-0.2in}
	\includegraphics[width=1.0\linewidth]{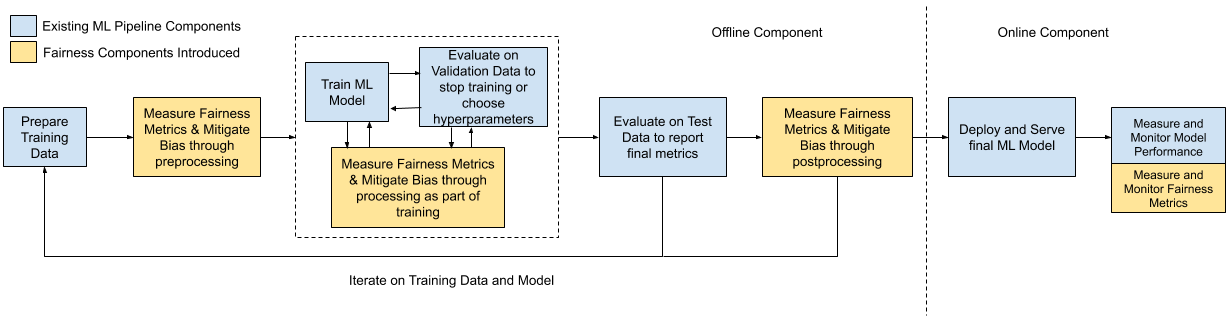}
	\vspace{-0.3in}
	\caption{High-level architecture of a ML training and serving system, showing how LiFT's bias measurement and mitigation components can be integrated.}
	\label{fig:fairnessSystemArch}
	\vspace{-0.1in}
\end{figure*}

\section{System Overview, Design, and Architecture}\label{sec:fairnessSystemArch}
We next give an overview of LiFT, our scalable framework for computing metrics for fairness in large-scale AI applications, and describe the system design and architecture.

LiFT comprises of bias measurement and mitigation components that can be intergrated into different stages of a ML training and serving system (Figure~\ref{fig:fairnessSystemArch}). It leverages Apache Spark~\cite{apacheSpark}, an open-source distributed general-purpose cluster-computing framework, to ensure that it can operate on datasets stored on distributed file systems as well as achieve data parallelism and fault tolerance. Utilizing Spark also provides compatibility with a variety of offline compute systems, ML frameworks, and cloud providers, for maximum flexibility.

LiFT is designed as a reusable library at its core, exposing several APIs and classes at various levels depending on how an ML practitioner wishes to interface with the system. Thus, users can easily leverage the library in ad-hoc analyses, deploy higher-level driver programs in offline workflows, or make use of a configuration-driven experience that integrates with LinkedIn's ML pipelines. The design of the library, its APIs, and configuration are described in \S\ref{sec:flexibleDesign}.

We next describe the high-level architecture of LiFT that enables it to integrate with LinkedIn's ML system, as well as the lower-level design that enables its flexibility and scalability.

\subsection{Integration with ML Pipelines}
A web-scale ML system can usually be divided into an offline model training component and an online model serving component. The offline component consists of workflows for preparing training datasets, training ML models (along with model validation and hyperparameter tuning), and for testing and benchmarking trained models. These workflows are often scheduled to run regularly so that new models can be trained on fresh data. The online serving system is responsible for identifying the correct model to be served, retrieving indexed features, computing predictions in a real-time manner, and measuring and monitoring model performance and business metrics on an ongoing basis. Here, we present the architecture of a fairness solution that integrates with almost all stages of the ML lifecycle (Figure~\ref{fig:fairnessSystemArch}):

\textbf{Before training:} This step deals with measuring metrics for representativeness and label distribution across different subgroups, as well as mitigation techniques such as~\cite{kamiran2012data}. By representativeness, we refer to measures for comparing the distribution of values of one or more protected attributes in the training data against a given desired (benchmark) distribution. Such measures are intended to inform the model developer about the extent to which the training data is representative across different subgroups, as desired for the application. We note that the desired distribution need not necessarily correspond to the population distribution. As an example, for applications such as gender determination from facial images or face recognition, improving model performance for minority groups may require oversampling from such groups. By label distribution, we refer to measures for comparing the label distributions across different subgroups (with each other), towards the goal of uncovering potential labeling biases within groups.
    
\textbf{During training:} This step enables the measurement of fairness metrics and bias mitigation during training, say, on each mini-batch during stochastic gradient descent or on the validation dataset during hyperparameter tuning. Such measurement can allow for choosing the right hyperparameters to enable the right balance of fairness and model performance. Furthermore, black-box mitigation techniques like \cite{agarwal2018reductions} or in-processing methods like~\cite{kamishima_2012} can be integrated as part of this module to obtain models that are optimal with respect to both performance and bias.
    
\textbf{After training:} This module enables measuring fairness metrics on the test dataset (for the final model) as well as post-processing mitigation methods like~\cite{geyik2019fairness}. Fairness measurement post model training can involve comparing predicted score  distributions across different subgroups (similar to measurements on training data), computing aggregate metrics of unfairness/inequality, or directly comparing performance metrics (such as AUROC, accuracy, PPV (precision), TPR (recall), and FPR) across different protected groups. In non-trivial cases of unequal prevalence rates, differences in model performance across subgroups can be expected due to the impossibility results~\cite{chouldechova2017fair}. Measuring multiple fairness metrics can thus be helpful to decide the appropriate tradeoffs, or to iterate on the training data and model altogether.
    
\textbf{Online serving:} The final component of the lifecycle involves measuring and monitoring fairness metrics on an ongoing basis. This step can not only be used for measuring these metrics during A/B tests, but also help with detecting issues like model drift with respect to fairness.

The above architecture is prescriptive, and the ability of a fairness system to integrate with an ML framework is dependent on the hooks exposed by that framework. To measure and mitigate bias before training, LiFT uses a hook exposed by LinkedIn's ML system meant for training data transformations. To enable post training measurement of fairness metrics, it utilizes the ML system's hooks for computing custom evaluation metrics. With a system that has hooks for custom metrics during training, one could use LiFT's existing capabilities to export the best model (thus far) based on fairness metrics, for example. LiFT is designed to be extensible so that additional components and functionality  for bias measurement and mitigation can be incorporated over time.

\subsection{Design for Flexibility}\label{sec:flexibleDesign}
For enabling use in exploratory settings as well in production workflows and ML pipelines, LiFT is designed as a library at its core, with wrappers and a configuration language meant for deployment (see Figure~\ref{fig:fairnessSoftwareDesign}).

\subsubsection{Customizability and Extensibility}
The library provides access to APIs that can be used directly to compute fairness metrics at various levels of granularity:
\begin{itemize}
    \item \textbf{High-level APIs:} These act as entry points for users wishing to integrate bias measurement and mitigation logic into their code, e.g., \textit{computeDatasetMetrics}, \textit{computeModelPerformanceMetrics}. These APIs bundle various low-level APIs and typically accept a parsed configuration instance.
    \item \textbf{Low-level APIs:} Users can also choose to utilize specific functionalities offered by LiFT, such as \textit{computePermutationTestMetrics} or \textit{computeJensenShannonDivergence}. This allow users to selectively work with only those capabilities they might care about.
\end{itemize}

\begin{figure}[htbp]
	\centering
	\includegraphics[width=1.0\linewidth]{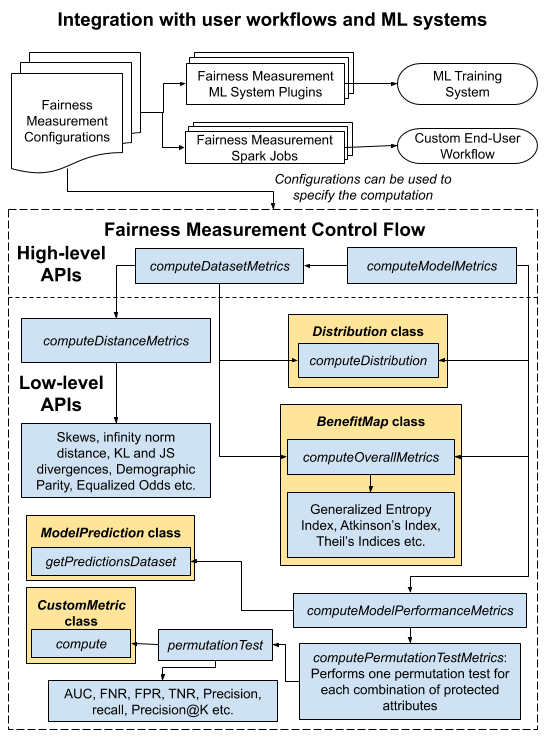}
	\caption{Design of LiFT and its interaction with external systems. The flowchart at the top shows how configuration-driven Spark jobs and ML plugins enable fairness metric computation in user workflows and ML systems respectively. At the bottom, the interaction between high and low level APIs and classes is shown.}
	\label{fig:fairnessSoftwareDesign}
	\vspace{-0.1in}
\end{figure}

Furthermore, key classes (shown in Figure~\ref{fig:fairnessSoftwareDesign}) can be extended to enable custom computation:
\begin{itemize}
    \item \textbf{\textit{Distribution} class}: Computes observed distributions of protected attributes, and calculates distance and divergence metrics with respect to other \textit{Distribution} instances.
    
    \item \textbf{\textit{BenefitMap} class}: Provides capabilities to capture benefit vectors and compute aggregate inequality metrics.
    
    \item \textbf{\textit{ModelPrediction} class}: Provides a standardized interface to compute model performance metrics, with support for data sampling and statistical tests (like permutation tests) to estimate if observed differences are statistically significant. Instances of this class can be used in collections for single system computation, or in \textit{Spark Datasets} for distributed computation.
    
    \item \textbf{\textit{CustomMetric} class:} Provides a specific interface that users can extend to define their own User Defined Functions (UDFs) that can be plugged into the fairness toolkit. Thus, while LiFT natively supports classification metrics, the \textit{CustomMetric} class enables defining metrics that allow it to be used in ranking scenarios as well.
\end{itemize}

\subsubsection{Configuration-Driven}
Once users have identified their metrics of interest and developed their own workflow, they may choose to deploy their own custom code that uses LiFT's APIs. However, the toolkit also provides a wrapper program that loads input datasets, joins them with other datasets containing protected attribute information, computes various fairness metrics, and writes out the report to a user-specified location. This program is available both as a Spark driver program, as well as a plugin for LinkedIn's ML framework. In both cases, the toolkit is driven by a user-specified configuration that allows for quick and easy deployment in production workflows, without requiring additional overheads of testing and verification. Shown below is an example configuration that specifies the input dataset of interest, the protected attribute dataset, the appropriate join keys, distance metrics to compute, and the aggregate inequality metrics to evaluate for each specified performance metric.

\begin{verbnobox}[\fontsize{8pt}{8pt}\selectfont]
'datasetPath': '/path/to/dataset',
'uidField': 'userIdField',
'labelField': 'datasetLabelField',
'scoreField': 'modelScoreField',
// Indicates if model scores are raw scores or probabilities
'scoreType': 'PROB', // raw scores passed through a sigmoid
'protectedAttributeField': 'protectedDatasetAttributeField',
'uidProtectedAttributeField': 'protectedDatasetJoinKey',
'protectedDatasetPath': '/path/to/protected/attributes',
'outputPath': '/path/to/output',
'distanceMetrics': ['DEMOGRAPHIC_PARITY', 'EQUALIZED_ODDS'],
// Use the following metrics to compute benefit vectors
'performanceBenefitMetrics': ['PRECISION', 'RECALL'],
// Specify benefit metrics to compute and their parameters
'overallMetrics': ['GENERALIZED_ENTROPY_INDEX': '0.5']
\end{verbnobox}

\subsection{Design for Scalability}
LiFT uses the following techniques to scale the computation of fairness metrics over billions of records:
\begin{itemize}
    \item Single job to load data files into in-memory, fault-tolerant and scalable data structures.
    \item Strategic caching of datasets and any pre-computation performed.
    \item Balancing distributed computation with single system execution to obtain a good mix of scalability and speed.
\end{itemize}

\begin{figure}[htbp]
	\centering
	\vspace{-0.05in}
	\includegraphics[width=1.0\linewidth]{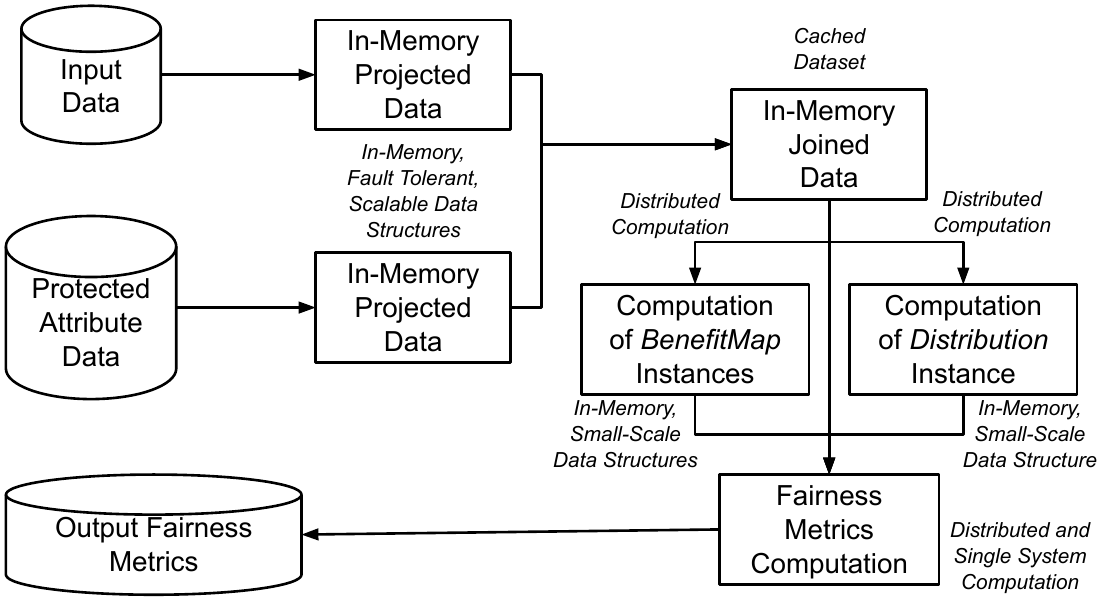}
	\vspace{-0.2in}
	\caption{Scalability Design in LiFT.}
	\label{fig:scalabilityDesign}
	\vspace{-0.05in}
\end{figure}

As shown in Figure~\ref{fig:scalabilityDesign}, prior to computing fairness metrics, the toolkit builds a virtual dataset comprised of only the primary key, labels and/or model predictions for the input dataset, and a second virtual dataset comprising the primary key and protected attributes for the dataset containing the protected attribute information. The datasets are loaded into Spark DataFrames~\cite{armbrust2015spark}, which are in-memory, fault-tolerant, and scalable data structures similar to a relational database. These are then preprocessed appropriately and joined to form the final virtual dataset of interest, which is cached to memory and disk to ensure quick downstream computation.

Fairness metric computations that involve distributions only work with \textit{Distribution} class instances. The creation of these instances results in the aggregation of the input dataset across a given set of dimensions and storing the resultant distribution in-memory, on a single system. This ensures that distribution comparisons can be performed quickly, avoiding the overheads associated with repeated distributed computation, and is efficient because these distributions are computed only once.

A similar idea is used for the \textit{BenefitMap} instances. Their creation results in the distributed computation of benefit metrics, with the resulting benefit vectors being stored in-memory on a single system. This ensures that subsequent computations of aggregate metrics of fairness (on these benefit vectors) are quick and efficient.

Users can choose to write custom metrics that operate on these precomputed \textit{Distribution} and \textit{BenefitMap} instances, thereby enabling quick single-system computation of these metrics. More involved metrics can instead operate on the cached virtual dataset, to ensure that the entire computation is distributed over several nodes. The Spark SQL Catalyst query optimizer and Spark Tungsten engine can leverage this end-to-end query plan to produce an optimized execution plan. This choice between single system and distributed computation is not only available to users, but is also used within LiFT for its natively supported metrics. While scalable to large datasets, making use of only distributed computation can encounter the following issues that affect its execution speed:
\begin{itemize}
    \item I/O overheads when loading data over several nodes,
    \item Network overheads, and 
    \item Query optimization overheads.
\end{itemize}

 For example, the computation of fairness metrics such as Demographic Parity or Equalized Odds (or metrics like TPR (True Positive Rate), FPR (False Positive Rate) and FNR (False Negative Rate), that are used to compute benefit vectors for aggregate metrics) revolve around the estimation of a `generalized confusion matrix'. A generalized confusion matrix uses expected values for true and false positives and negatives, thereby requiring that the model only assign a $P(y=1|x)$ value for each data point $x$, rather than needing them to pick a classification threshold as well. The computation of such a matrix can be distributed easily, with the map operation counting the true and false positives and negatives, and the reducer summing up these counts.
 
However, the non-parametric statistical tests~\cite{diciccio2020fairness} that LiFT supports involve computations performed over several trials, to obtain a distribution of the test statistic. Attempting to distribute these trials over several nodes led to a significant slowdown in the computation time due to Spark's query plan optimization process, whose runtime is non-linear in the number of sub-queries. Furthermore, these tests achieved sufficient statistical power with about $100,000$ data points, indicating that we did not have to distribute the computation of each trial, and that retrieving a random sample of datapoints from the distributed virtual dataset was sufficient. Since each trial of the statistical test involved random sampling or shuffling of the data, if it were evaluated using a distributed computing framework, we would have triggered several shuffle steps and created massive network overhead, thereby slowing down computation even more. Thus, we used Spark to project only the desired columns and sample the large dataset efficiently, and utilized single system computation to perform the actual statistical test.

Finally, if the dataset is small enough or the cardinality of the protected attributes is small, it might be beneficial to stick to single system execution for the final metrics computation to avoid slowdowns due to repeated disk reads or other distributed overheads.
\section{Metrics Measured}\label{sec:metrics}
Currently, LiFT supports the measurement of fairness metrics pre- and post- training, as well as the mitigation of bias using a variant of the preprocessing technique described in~\cite{kamiran2012data}. The decision to support measurement during these two stages was primarily driven by the availability of hooks into LinkedIn's ML system, to ensure easy integration and adoption. Nevertheless, this does not preclude capabilities such as measuring metrics during training. LiFT is also designed to be extensible, enabling users to plug-in their own metrics using User Defined Functions (UDFs), incorporate individual modules as part of data validation systems~\cite{swami2020data}, or enhance its capabilities to address more measurement and mitigation strategies during various stages of the ML lifecycle.

\subsection{Fairness Metrics for Training Data}
To measure the representativeness and label distribution of training data, LiFT supports various distribution and divergence metrics:
\begin{enumerate}
    \item \textbf{KL and JS Divergences:} These metrics can be used to compute a weighted sum of logarithmic differences between a reference distribution and the observed distribution of protected attributes. Let $P$ be the desired distribution and $Q$ be the observed distribution. Then, the KL divergence is
    $KL(P||Q) = \sum_x P(x)\log\left(\frac{P(x)}{Q(x)}\right)$.
    Thus, a uniform reference distribution would weight all differences equally, while using the protected attribute distribution of a population would ensure that differences are weighted by the population density of the protected groups. The JS Divergence is similar to the KL Divergence, but captures a symmetric difference between the reference and target distributions. If $M$ is the average distribution of $P$ and $Q$, the JS Divergence is the mean of the KL Divergences from $M$ to $P$ and from $M$ to $Q$.
    
    \item \textbf{L-p Norm and Total Variation Distances:} For $p \geq 1$, $L_p(P,Q) = \left(\sum_x \left|P(x) - Q(x)\right|^{p}\right)^{1/p}$.
    The Total Variation Distance is equal to half the $L_1$ distance, and is the largest possible difference between the probabilities that the two distributions can assign to the same event.
    
    \item \textbf{Demographic Parity:} The above metrics capture representativeness with respect to a reference distribution. We make use of the idea of Demographic Parity to measure label distribution differences between protected groups. Demographic Parity typically requires that $P(\hat{Y} = 1| G=g_1) = P(\hat{Y} = 1| G=g_2) \;\forall g_1,g_2\in G$,
    where $\hat{Y}$ is the model's prediction and $G$ is the protected attribute. Repurposing this for training data, we measure
    $\delta_{DP}(g_1, g_2) = | P(Y = 1| G=g_1) - P(Y = 1| G=g_2)|$
    where $Y$ is the label field in the training data. We thus desire $\delta_{DP}(g_1,g_2) = 0\; \forall g_1, g_2 \in G$ to ensure that $G$ and $Y$ are statistically independent in the training data.
\end{enumerate}

\subsection{Fairness Metrics Post Model Training}
LiFT supports the following categories of metrics for measuring fairness of the final model:
\begin{enumerate}
   \item \textbf{Distance and Divergence Metrics:} These metrics are similar to those computed for the training data, but make use of the model predictions instead of the labels. Note that this class of metrics includes a measure of the extent to which Demographic Parity is violated, i.e., $| P(\hat{Y} = 1| G=g_1) - P(\hat{Y} = 1| G=g_2)|$.
   We also compute additional metrics such as Equalized Odds, which captures the differences in True and False Positive Rates for different protected groups~\cite{hardt_2016}: $\delta_{EO}(y, g_1, g_2) = | P(\hat{Y} = 1| Y=y, G=g_1) - P(\hat{Y} = 1| Y=y, G=g_2)|$.
    
    \item \textbf{Aggregate Fairness Metrics:} These are inequality measures such as the Generalized Entropy Index~\cite{speicher2018unified} (and related metrics like Theil's L and T Indices) computed over different choices of benefit vectors. A benefit vector is a vector whose elements (benefit values) capture some notion of benefit (or model performance) for each entity across which inequality is to be measured. Benefit values can be defined at the individual level or computed for each protected group, depending on whether we are measuring individual or group notions of fairness. An inequality measure $I$ maps a benefit vector \textbf{b} to a non-negative real number $I(\textbf{b})$, with \textbf{b} being considered more fair than \textbf{b'} if and only if $I(\textbf{b}) < I(\textbf{b'})$.
   
    \item \textbf{Differences in Model Performance:} We also compute statistically significant differences in performance metrics for each pair of subgroups using permutation tests~\cite{diciccio2020fairness, good2000permutation, ojala2009}. FairTest~\cite{tramer2015fairtest} is the only other toolkit with such a test. However, the implementation there is not applicable in all scenarios, while LiFT's version supports custom metrics~\cite{diciccio2020fairness}.
\end{enumerate}    
\section{Deployment Results}\label{sec:results}
We deployed LiFT as part of three web-scale ML pipelines at LinkedIn to measure fairness metrics as part of regular model training. These pipelines make use of internal datasets to learn models that power various products. Two of these were deployed as Spark jobs in custom workflows, while the third was deployed as a plugin into LinkedIn's ML framework:

\begin{itemize}
    \item Dataset $D_1$ consisted of about 28M records with around 50 continuous features, used to train a classifier evaluated on recall at a fixed value of precision. The model is used to keep LinkedIn members safe on the website.
    \item Dataset $D_2$ consisted of about 3.5M records with approximately 600 features, used to train another classifier evaluated on recall at a specific precision value. This model too keeps LinkedIn members safe.
    \item Dataset $D_3$ consisted of about 23M records with around 130 continuous features, used to train a ranking model evaluated by its AUC. This model is used by the Jobs platform.
\end{itemize}

{\noindent \bf Public Dataset}: To ensure reproducibility, we also evaluated our solution on the Adult dataset ($D_4$) from the UCI ML Repository \cite{kohavi1996scaling}. It consists of $48842$ records, which we split into a $70\%-15\%-15\%$ train-validation-test split. We made use of all the features available except gender, race, and the `final weight' feature, and one-hot-encoded categorical features. We then trained a logistic regression model with L2 regularization, resulting in a final model with an AUC of $0.909$. To ensure that LiFT's performance could be compared across all 4 datasets, we oversampled the test dataset ($\sim7400$ records) to generate a final dataset of about $10M$ records.

In this section, we present results that validate LiFT's flexibility and scalability. We discuss the experiences of our deployments, the challenges faced and lessons learned in the next section.

\subsection{Flexibility}
Since LiFT provides a Spark driver program as well as a plugin for the in-house ML framework, integration with production pipelines was extremely straightforward, with only a configuration file needed to measure metrics of interest. All three pipelines used a combination of gender (male, female and unknown) and country as their protected attributes; however $D_1$ and $D_2$ used multiple fairness metrics with respect to precision and recall, while $D_3$ focused on measuring fairness using permutation tests for AUC. The ease of setup, deployment, and measurement of bias despite the differences in the combinations of workflows and fairness requirements validated LiFT's flexibility.

\subsection{Scalability} \label{sec:scalability}
To validate the scalability of our framework, we made use of the same configuration across all four datasets to ensure that the results can be compared:
\begin{verbnobox}[\fontsize{8pt}{8pt}\selectfont]
protectedAttributeField: "gender",
distanceMetrics: "INF_NORM_DIST,TOTAL_VAR_DIST,JS_DIVERGENCE,
                  KL_DIVERGENCE,DEMOGRAPHIC_PARITY,
                  EQUALIZED_ODDS",
overallMetrics: "GENERALIZED_ENTROPY_INDEX=0.5,
                 THEIL_L_INDEX=,THEIL_T_INDEX=",
performanceBenefitMetrics: "AUC,PRECISION,RECALL,FPR"
\end{verbnobox}

As shown above, we measured six distance and divergence related metrics using a uniform gender-label distribution as the reference (wherever necessary). We also computed three aggregate metrics on four benefit vectors, each of which was comprised of model performance metrics for each protected group. Thus, the same set of eighteen metrics were computed across all four datasets, allowing us to measure the scalability of the implemented solution. 

\begin{figure}[htbp]
    \vspace{-0.05in}
	\centering
	\includegraphics[width=0.8\linewidth]{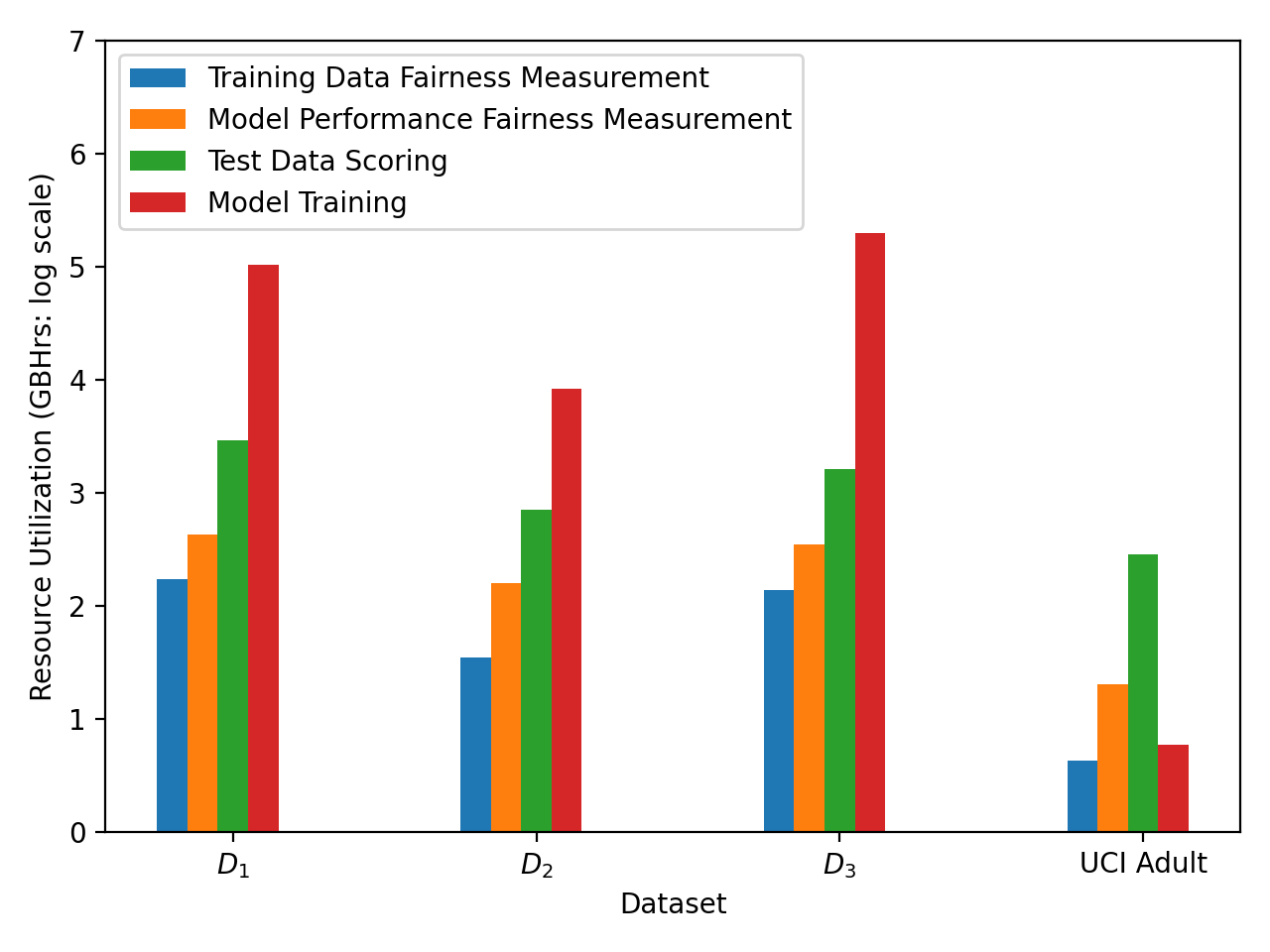}
	\vspace{-0.2in}
	\caption{Resource utilization (in log scale) for model training, scoring, and fairness metrics computation for 4 large-scale datasets.}
	\label{fig:fairness_perf}
	\vspace{-0.05in}
\end{figure}

Figure \ref{fig:fairness_perf} summarizes the results, depicting the resource utilization of various stages measured in GB-Hours\footnote{It is the product of the memory used, the number of nodes, and the time taken, and ensures that a partial picture is not presented where time is traded off for memory, or vice versa. GB-Hours and related measures are also used by cloud computing vendors to estimate usage and determine billing.}. We have also presented the data for model training and test data scoring stages, for reference. Due to the stark differences in resource consumption, we represent the Y-axis in the log scale. Note that for dataset $D_4$ (the Adult dataset), training was done on about $34000$ data points but the other stages were performed on the oversampled test dataset consisting of $10M$ records.

We see that the fairness measurement stages for both training data and model performance are able to scale to datasets of different sizes. Furthermore, they consume an order of magnitude fewer resources than test data scoring, with the distance computations for the training data being the cheapest to evaluate. That is, their computation is at least as efficient as that of model scoring. In our experiments, we kept the number of nodes and their memory constant, thereby ensuring that resource utilization (measured in GB-Hours) is directly proportional to the time taken. Then, we can clearly see that fairness metrics computation runs several orders of magnitude faster than model training, thereby adding a negligible amount of overhead to the overall model development process.

Since none of the datasets $D_i$ make use of protected attributes, about half the fairness metric computation time is actually spent in joining them with an independent dataset containing these attributes. Since these metric measurements are computed only using the labels, scores, and protected attributes of each data point, their runtimes are largely independent of the number of features present in the dataset. Furthermore, the computation is also model agnostic in terms of its type, complexity, or training time.

\subsection{Effectiveness of Metrics Measured}
While LiFT does not propose any new fairness measures, we nevertheless address the effectiveness of the supported metrics in this section. In our experiments, we measured six distance metrics and twelve aggregate benefit metrics for three pairs of gender groups. Shown below is an excerpt of fairness metrics computed for the UCI Adult dataset (parameters omitted for brevity):

\begin{center}
\begin{tabular}{ |c|c|} 
 \hline
 Generalized Entropy Index (GEI) for TPR & 0.0028 \\ 
  \hline
 Generalized Entropy Index (GEI) for FPR & 0.0908 \\ 
  \hline
 Equalized Odds (EO) for label = 1 & 0.0821 \\ 
  \hline
 Equalized Odds (EO) for label = 0 & 0.0998 \\ 
  \hline
 p-value (permutation tests for TPR and FPR) & 0.0 \\
 \hline
 $P(\hat{Y} = 1 | Y = 1, g = \text{`Female'})$ & 0.5111 \\
 \hline
 $P(\hat{Y} = 1 | Y = 1, g = \text{`Male'})$ & 0.5932 \\
 \hline
 $P(\hat{Y} = 1 | Y = 0, g = \text{`Female'})$ & 0.0706 \\
 \hline
 $P(\hat{Y} = 1 | Y = 0, g = \text{`Male'})$ & 0.1704 \\
 \hline
\end{tabular}
\end{center}

We see that GEI and EO are not easily interpretable / actionable (with respect to the degree of `unfairness'), while the permutation test clearly indicates that these differences are statistically significant. However, GEI and EO values can be charted over time, and changes can be detected easily, while the permutation test provides a binary result.

Also, while the permutation test indicates if a difference is statistically significant, it is up to the end-user to ultimately decide if this difference is worth acting upon. 
For example, the TPR is about $14\%$ lower for the `Female' group, while the FPR is about $58\%$ lower for the `Female' group. 
These differences are large enough to be statistically significant, causing the permutation test to reject in both cases. However, the GEI for TPR is much lower than that for FPR because the TPR values for the two groups are more similar than their corresponding FPR values.

Thus, no single metric can be deemed the most effective; rather a subset must be inspected to understand the larger picture and make appropriate decisions pertaining to bias. In practice, users should choose an appropriate set of metrics based on business and fairness objectives and properties of their data, so that focus can be placed only on those signals of interest. Depending on the context, some metrics might be redundant while others might provide conflicting signals (at first glance), and this insight can help narrow down the metrics to track on a regular basis.
\section{Deployment Impact}\label{sec:impact}
\subsection{Insights Gained from Deployments}
For $D_1$, our initial observation was that the model was unbiased between genders globally and on a per-country basis, except for a bias against the `unknown' gender in one country. Upon closer inspection, we discovered that our gender coverage was poor for that country, with very few `male' and `female' data points and the rest being assigned to the `unknown' bucket. This made us realize that (a) our metrics are only as good as our data, and (b) small sample sizes lead to unreliable estimates with high variance.

For $D_2$, EO showed us that the model was biased with respect to recall (TPR), but unbiased with respect to FPR. This was however due to the fact that only two of \{Precision, Recall, FPR\} can be equal in populations with different prevalence rates $p$ (percentage of positively labeled data points), since balancing any two requires that the third be imbalanced: $FPR = \frac{p}{(1-p)} \cdot \frac{(1-PPV)}{PPV} \cdot TPR$. Given that the prevalence rates varied across gender and that the model required a fixed high value of precision to be achieved, this `impossibility result' caused an imbalance in recall while maintaining equality in extremely low FPR values. Thus, there can be conflicting notions of fairness (unbiased precision versus unbiased recall), and a consensus must be reached with multiple stakeholders to make a final decision. In this case, unbiased precision would ensure that the related member experience would be great, while the unbiased recall was not an issue since false negatives were being addressed by a human-in-the-loop component.

Regarding $D_3$, we observed that the model was unbiased with respect to AUC. While this model was fair in the sense that it ranks relevant results higher than irrelevant ones in a similar manner across genders and countries, it brought to light that there can be several more notions of `fair ranking', and a plethora of potential attributes that we could measure bias against.

\subsection{Challenges Faced and Lessons Learned}
Based on our experience in deploying fairness measurement solutions in web-scale, production ML systems, we highlight key challenges that must be addressed for widespread adoption of fairness in the ML lifecycle, and lessons learned through the development and deployment of the LinkedIn Fairness Toolkit (LiFT). In light of the socio-technical nature of many ML applications, we emphasize that decisions pertaining to fairness need to be guided by social, ethical, and legal dimensions, and based on inputs from a diverse set of key stakeholders.

Adoption of a fairness toolkit is easiest when it integrates with an ML service, making the entire process frictionless for the model developer. Production settings also require that the adopted toolkit scale to large datasets. Results should be easily interpretable since ML developers and stakeholders might otherwise be overwhelmed by the amount of fairness metrics, their definitions and potentially conflicting signals. Support for custom metrics (through UDFs) is also necessary for adoption, since different products and ML models have different sets of metrics that they care about, and supporting different measures can get unwieldy pretty quickly.
    
Focusing on a few fairness metrics and protected attributes greatly simplifies the effort needed to measure and mitigate bias. Furthermore, identifying other key segments such as countries or educational qualifications also helps monitor fairness within these cohorts. Different applications have different requirements and hence, different fairness needs. For example, a model that identifies bad actors in an ecosystem might want to ensure extremely high and equal precision for all protected groups, but might be relaxed about its recall requirements. The fairness notion for a ranking system could require the top retrieved results to be representative of the candidate pool~\cite{geyik2019fairness}. The impossibility results also imply that not all fairness requirements can be simultaneously achieved. We thus need to consider ML fairness in the context of specific definitions and requirements, and make note of any caveats or assumptions upfront, using model cards for example~\cite{mitchell2019model}.
    
Statistical notions of fairness are necessary to enable bias measurement over large noisy datasets, providing measures such as p-values, confidence intervals, and standard errors. Statistical fairness concepts will not only make bias decision thresholds clearer, but also help account for uncertainty in protected attribute values.
    
To ensure that ML models are unbiased with respect to certain notions of fairness, we may need access to users' protected attributes. Such information may not be easily available and users may need to volunteer to provide this data. Furthermore, given access to this data, we need to ensure that this information is only used for bias measurement and mitigation purposes, and that data access is otherwise tightly controlled. In the absence of user-provided information regarding protected attributes, there may be a need to infer such attributes. However, this introduces an element of uncertainty in the data quality, which must be accounted for in our estimates of bias and mitigation strategies. Also, there might be cases where attribute inference might be impossible or ethically inappropriate.
\section{Related Work} \label{sec:relatedwork}
Existing fairness toolkits are designed to be run either as standalone instances on a single machine, or with a specific cloud provider / ML framework. For example, IBM's open sourced AI Fairness 360 tool~\cite{bellamy2018ai} is designed for execution on a single machine (over appropriately sized datasets), making use of ML frameworks like `scikit-learn'~\cite{pedregosa2011scikit}. While we are unaware if the IBM Cloud offering of this library is a distributed variant, its tight integration with this platform nevertheless requires that offline workflows and ML pipelines use IBM Cloud. Microsoft AzureML Interpret is similar in this regard, requiring the adoption of Azure. Google's What-If Tool works only with TensorFlow~\cite{abadi2016tensorflow}. Thus, the deployment of these fairness toolkits is hard without the wholesale adoption of these cloud providers / ML frameworks. LiFT however is a more flexible system, enabling deployments in a wider variety of settings.

Publicly available fairness toolkits such as Aequitas~\cite{saleiro2018aequitas}, Audit-AI~\cite{auditAI}, FairLearn~\cite{agarwal2018reductions}, FairML~\cite{adebayo2016fairml}, Fairness Comparison~\cite{friedler2019comparative}, Fairness Measures~\cite{fairnessMeasures}, FairTest~\cite{tramer2015fairtest}, Themis~\cite{galhotra2017fairness}, and Themis-ML~\cite{bantilan2018themis} provide access to higher and lower-level APIs but seem to be designed for single system execution only, thereby being constrained to operate on relatively smaller datasets. As a result, despite their comprehensive suite of metrics and mitigation strategies, these toolkits cannot be directly deployed in production settings. On the other hand, LiFT is designed to leverage distributed computation where possible, and integrates easily with offline workflows.

Due to the above solutions being single system based Python libraries, LiFT cannot be compared against them in the context of scalability and efficiency. One possible approach is to materialize large distributed datasets onto a single node to allow these libraries to compute fairness metrics. However, this can result in long processing times or failure to compute metrics altogether. Furthermore, requiring that LiFT operate on a single node (for these comparisons) would only seek to compare underlying technologies and not the libraries themselves. It is also not straightforward to extend existing single machine toolkits to enable distributed computation, since this would require either a complete distributed-first overhaul of their libraries, or necessitate an under-the-hood translation of their compute logic to run on several nodes in a parallel manner.

Finally, while most related work aims to provide a comprehensive set of fairness measures, LiFT primarily focuses on providing a framework that is flexible, scalable and easy to integrate into offline workflows. However, these goals do not imply that only metrics amenable to distributed computation are part of this library, nor does it preclude the addition of new measurement techniques. It is important to note that LiFT's scalability comes at the cost of distributed computation overhead for smaller datasets. This can nevertheless be reduced by tuning the amount of resources and the number of nodes required. No such tuning can be done to scale single system computation.
\section{Conclusion and Future Work}\label{sec:conclusion}
Considering the importance of measuring and mitigating algorithmic bias in large-scale ML based applications, we presented the LinkedIn Fairness Toolkit (LiFT), a system for scalable and flexible computation of fairness metrics during different stages of the ML lifecycle. We highlighted the key requirements of such a system, described its overall design as well as individual components, and highlighted the practical challenges faced and lessons learned during deployment. We are in the process of \textit{open-sourcing} our library towards wider use by researchers and practitioners.

We discuss a few directions for future work based on our experiences. Considering that typical ML based products/services consist of several ML models and non-ML components composed together in different stages, a fruitful direction is to explore approaches for measuring fairness for the product/service as a whole. Another potential direction is to explore notions and definitions of fairness for `human-in-the-loop' / `algorithm-in-the-loop' systems~\cite{green2019disparate}. Whether collected from users or inferred, there is the possibility of errors in the values of protected attributes, or values being missing altogether. Thus, another area of research would involve accounting for these uncertainties in estimates of bias, or during bias mitigation.

From a system design perspective, one direction for future work involves revisiting existing measurement and mitigation strategies to design them with scalability in mind, thereby enabling easy adoption in production. Yet another direction involves extending distributed measurement and mitigation solutions to nearline and streaming settings, to measure fairness metrics on an ongoing basis and apply bias mitigation strategies in a reactive manner.

Successful and broad adoption of fairness tools such as LiFT is contingent on collaborating with a diverse set of key stakeholders (spanning product, legal, policy, PR, engineering, AI/ML, and other teams) and building consensus on the desired notions of bias and fairness. A constructive direction is to further understand the societal perspectives and needs of practitioners (e.g., \cite{srivastava2019mathematical, grgic2018human, holstein2019improving, beutel2019putting}), and to investigate the role of bias measurement and mitigation tools as part of such a holistic approach.

{
\bibliographystyle{abbrv}
\bibliography{paper}
}

\end{document}